# LAF-Based Evaluation and UTTL-Based Learning Strategies with MIATTs

Yongquan Yang[1*]

[1]Institute of Sciences for AI, Chengdu, Sichuan, China

[*]Corresponding author (Email: remy_yang@foxmail.com or yongquan.yang@sciences4ai.com, ORCiD: 0000-0002-3965-4816)

In many real-world machine learning (ML) applications, the true target cannot be precisely defined due to ambiguity or subjectivity information. To address this challenge, under the assumption that the true target for a given ML task is not assumed to exist objectively in the real world, the EL-MIATTs (Evaluation and Learning with Multiple Inaccurate True Targets) framework has been proposed. Bridging theory and practice in implementing EL-MIATTs, in this paper, we develop two complementary mechanisms: LAF (Logical Assessment Formula)-based evaluation algorithms and UTTL (Undefinable True Target Learning)-based learning strategies with MIATTs, which together enable logically coherent and practically feasible modeling under uncertain supervision. We first analyze task-specific MIATTs, examining how their coverage and diversity determine its structural property and influence downstream evaluation and learning. Based on this understanding, we formulate LAF-grounded evaluation algorithms that operate either on original MIATTs or on ternary targets synthesized from them, balancing interpretability, soundness, and completeness. For model training, we introduce UTTL-grounded learning strategies using Dice and Cross-Entropy loss functions, comparing per-target and aggregated optimization schemes. We also discuss how the integration of LAF and UTTL bridges the gap between logical semantics and statistical optimization. Together, these components provide a coherent pathway for implementing EL-MIATTs, offering a principled foundation for developing ML systems in scenarios where the notion of "ground truth" is inherently uncertain. An application of this work's results is presented as part of the study available at https://www.qeios.com/read/EZWLSN.

## 1. Introduction

Modern machine learning (ML) systems are increasingly deployed in real-world environments where the notion of a single, precisely defined true target is often ill-posed or even nonexistent. In many practical domains, such as medical diagnosis, social behavior analysis, and open-world perception, ground truth labels arise from subjective, incomplete, or conflicting human or model-generated sources. These conditions challenge the classical assumption of a deterministic and accurate true target (ATT), which underpins most conventional evaluation [1–5] and learning [6–10] paradigms. To bridge this theoretical–practical gap, under the explicitly posited assumption that the true target for a given ML task is not assumed to exist as a well-defined object in the real world, the EL-MIATTs (Evaluation and Learning with Multiple Inaccurate True Targets) framework has been proposed as a means to model, analyze, and utilize imperfect yet informative approximations of the underlying truth [11].

While the theoretical formulation of EL-MIATTs establishes a principled basis for addressing epistemic uncertainty, its practical implementation, which extends beyond the generation and assessment of MIATTs [12], critically depends on two complementary mechanisms that operationalize the framework in real-world ML tasks: LAF (Logical Assessment Formula [13])-based evaluation and UTTL (Undefinable True Target Learning [14])-based learning with MIATTs. Accordingly, this paper presents applicable LAF-grounded evaluation algorithms and UTTL-grounded learning strategies as the operational realization of these two mechanisms within the EL-MIATTs framework. The former enables logically consistent assessment of models when multiple partially correct targets coexist; the latter allows model training to proceed effectively even when the true target is fundamentally undefinable. Together, these mechanisms transform EL-MIATTs from a conceptual framework into a deployable methodology for complex, real-world machine learning tasks. Their integration is thus crucial for achieving both theoretical rigor and practical utility in uncertain supervision settings. Preliminaries related to EL-MIATTs framework are provided in Section2.

A central component of this study is the analysis of task-specific MIATTs, which are generated from diverse task-related AI models (AIM) retrievable from real-world resources [12]. Each MIATT represents a partial, probabilistic, or context-specific approximation of the latent true target. The quality of a task-specific MIATTs set can be characterized by its coverage (mean of PartialRepresentation) and diversity (1 − Redundancy), together determining its representational fidelity. High-quality MIATTs achieve a balance between completeness and consistency, covering the semantic scope of the true target while minimizing redundant or contradictory information. As this structural property is essential for designing reliable evaluation and learning processes under EL-MIATTs, we also analyze the downstream influence of task-specific MIATTs on evaluation and learning. A comprehensive analysis of the structural property of task-specific MIATTs and its downstream influence on evaluation and learning is presented in Section 3.

LAF [13] provides a theoretical and algorithmic foundation for evaluating predictive models with MIATTs. Building upon its principle, we extend classical logic-based and fuzzy operations [15–19] to accommodate evaluation with MIATTs, thereby supporting both parallel multi-perspective and ternary synthesized evaluation schemes. In the parallel multi-perspective evaluation scheme, each MIATT preserves its individual partial truth and contributes to the overall assessment through logical aggregation operations such as conjunction, disjunction, t-norm, and t-conorm. This approach maintains detailed information and enables fine-grained interpretability. In contrast, the ternary synthesized evaluation scheme compresses the MIATTs set into a single three-valued representation $\{0, 0.5, 1\}$, facilitating unified scoring and computational simplicity at the expense of some informational granularity. Together, these LAF-based evaluation algorithms achieve a balance between logical completeness and practical interpretability, approximating conventional ATT-based evaluation in complex, ill-defined tasks while reflecting intrinsic ambiguity in simpler or underdefined scenarios. Further methodological details regarding these two LAF-based evaluation algorithms are presented in Section 4.

Complementing LAF, UTTL [14] addresses the challenge of model training when the true target is uncertain or inherently undefined. UTTL can be operationalized by treating MIATTs as multiple weakly reliable surrogates of the ground truth, enabling the learning process to proceed through multi-target optimization [14]. Building on this principle, two main strategies can be derived: (1) Per-target then Aggregate, which computes loss functions (e.g., Dice [20] or Cross Entropy [21]) for each MIATT individually before aggregating the results; and (2) Aggregate then Single Loss, which first synthesizes a composite target and then computes a single loss against it. Depending on the loss function used for optimization, these two strategies exhibit different theoretical behaviors and learning biases—one emphasizing robustness to diversity, the other promoting consistency and stability. Collectively, they establish a flexible paradigm for learning under epistemic uncertainty, fostering the development of adaptive and explainable ML systems. Further theoretical analyses and practical guidelines for applying these two UTTL-based learning strategies are presented in Section 5.

Synthesizing the structural property of task-specific MIATTs and its downstream influence on evaluation and learning, and the theoretical implications and practical insights derived from LAF-based evaluation and UTTL-based learning, we further discuss how the integration of LAF and UTTL bridges the gap between logical semantics and statistical optimization. This discussion examines the inherent trade-offs among coverage, consistency, and interpretability within multi-valued logic systems, extending the EL-MIATTs framework toward paraconsistent reasoning and adaptive weighting mechanisms. Comprehensive discussion and analytical results are presented in Section 6.

In summary, this work advances the theoretical and practical development of the EL-MIATTs framework by introducing concrete algorithmic and strategic realizations for both evaluation and learning with MIATTs, establishing a foundation for reliable, interpretable, and uncertainty-aware machine learning. The key contributions of this work are summarized as follows:

- We analyze the structural properties of task-specific MIATTs using coverage- and diversity-based indicators and discuss its downstream influence on evaluation and learning.
- We propose two LAF-grounded algorithms for evaluation with MIATTs, supporting both parallel multi-perspective and ternary synthesized schemes.
- We develop and summarize two UTTL-grounded learning strategies for model training with MIATTs, based on Dice and Cross-Entropy losses.
- We further discuss these three contributions to bring out the integration of LAF-based evaluation with and UTTL-based learning with MIATTs, extending the EL-MIATTs framework to bridge logical semantics with statistical optimization while enhancing robustness, interpretability, and adaptability in uncertainty-aware learning.

The remainder of this paper is structured as follows: Section 2 introduces the preliminaries related to EL-MIATTs. Section 3 analyzes the structural properties of task-specific MIATTs with respect to their assessment indicators. Section 4 presents the two proposed LAF-grounded algorithms for evaluation with MIATTs. Section 5 describes two UTTL-grounded learning strategies for model training with MIATTs. Section 6 discusses the integration of LAF-based evaluation and UTTL-based learning to extend the EL-MIATTs framework. Finally, Section 7 concludes the paper, highlighting limitations and directions for future research.

## 2. Preliminary

In this section, building on previous studies [11, 12], we briefly introduce the definition and core concept of MIATTs, their task-specific generation and evaluation processes, as well as LAF [13] and UTTL [14] for evaluation and learning of predictive models with MIATTS.

### 2.1 Definition and essence of MIATTs

Building on the core premise that the true target for a given machine learning task is not assumed to exist as a well-defined object in the real world, the notion of MIATTs can be introduced as:

Let $t^*$ be the underlying (possibly undefinable) true target and $SF(t^*)$ its set of semantic facts. A MIATTs set is $MIATTs = \{t_n^* | n \in \{1, \cdots, N\}, N \geq 2\}$, where each $t_n^*$ satisfies

$$SF(t_n^*) \subset SF(t^*) \text{ and } \bigcup_{n=1}^{N} SF(t_n^*) \subseteq SF(t^*). \tag{1}$$

In other words, each $t_n^*$ reflects only part of the semantics of $t^*$, while the collection as a whole provides a broader approximation of it.

Fundamentally, MIATTs capture the insight that supervision in real-world machine learning is inherently partial and noisy [11]. While any single inaccurate true target conveys just a fragment of the underlying semantic structure, the set collectively reconstructs it with wider coverage. This reframing shifts supervision away from a strict single-target view toward a distributional, multi-perspective paradigm, thereby laying a principled basis for the EL-MIATTs framework to support robust evaluation and learning under uncertainty.

### 2.2 Generation and assessment of task-specific MIATTs

Based on this definition and essence of MIATTs, logic-driven algorithms have been proposed for generation and assessment of MIATTS [12]. In this section, we summarize two simplified logic-driven solutions using retrievable real-world resources for generation and assessment of task-specific MIATTs.

#### 2.2.1 Generation of task-specific MIATTs

For specific tasks, existing resources—such as accumulated task-specific datasets, pretrained models, or related large AI models—can be readily utilized. By leveraging these resources, we construct a set of task-specific AI models (AIM), integrating both prior task-specific models and relevant large AI models. This AIM set maps each instance to multiple predicted true targets corresponding to its underlying ground truth, collectively forming the generated MIATTs. Let $AIM = \{p_1, p_2, \ldots, p_N\}$ denote the set of predictive models and tools derived from these resources. The stepwise procedure of this approach [12] is outlined as follows:

**Input:**
- Raw data instances $I$.
- A set of AI models $AIM = \{p_1, p_2, \ldots, p_t\}$ for mapping an instance into predicted multiple true targets for the underlying true target $t^*$.

**Step 1: Prediction of multiple potential true targets**
- Predict multiple potential true targets for $I$ with $AIM$:
$$MIATTs = AIM(I) = \{p_1(I), p_2(I), \ldots, p_N(I)\} = \{t_n^* | n \in \{1, \cdots, N\}\}, \ N \geq 2. \quad (2)$$

**Output:**
- Each instance in $I$ is assigned an MIATTs set $\{t_n^*\}_{n=1}^{N \geq 2}$, each being a partial but informative approximation of the corresponding underlying true target $t^*$.

### 2.2.2 Assessment of task-specific MIATTs

For a specific task, once an MIATTs set is generated from task-specific AIM retrieved from real-world resources, a probable true target can be derived to approximate the underlying ground truth. Since each IATT in the MIATTs set covers a partial yet reliable aspect of the true target, integrating these partial coverages yields a synthesized target that captures the essential semantics of the ground truth. This summarized probable target then serves as a reference for evaluating the MIATTs set itself, ensuring a self-consistent and task-adaptive assessment process. The stepwise procedure [12] is outlined as follows:

**Input:**
- An MIATTs set $\{t_n^*\}_{n=1}^{N \geq 2}$ generated by task-specific AIM.

**Step 1: Approximate probable true target**
$$\breve{t}^* = mean(\{t_n^*\}_{n=1}^{N \geq 2}). \quad (3)$$

**Step 2: Represent IATT as Boolean vectors**
- Represent each IATT $t_n^*$ as a Boolean vector $v_n \in \{0,1\}^m$, where:
$$v_n[i] = \begin{cases} 1, & if \ abs\left(t_n^*(i) - \breve{t}^*(i)\right) < \delta \\ 0, & otherwise \end{cases}. \quad (4)$$

**Step 3: Assess partial representation (Per-IATT quality)**
- For each IATT $t_n^*$:
$$PartialRepresentation(t_n^*) = \frac{\sum_i v_n[i]}{m}. \quad (5)$$

**Step 4: Assess redundancy / diversity**
- Compute pairwise intersections (logical AND) between MIATTs:
$$Intersection(t_j^*, t_k^*) = v_j \wedge v_k.$$
- Measure redundancy ratio:
$$Redundancy = \frac{\sum_{j<k} |v_j \wedge v_k|}{\sum_j |v_j|}. \quad (6)$$

**Step 5: Overall quality score**
- Combine the metrics into an aggregate score:
$$Q_{MIATTs} = \alpha \cdot mean(PartialRepresentation) - \gamma \cdot Redundancy. \quad (7)$$

**Output:**
- The computed $Q_{MIATTs}$ for assessing the overall quality score of the MIATTs set.

## 2.3 LAF and UTTL

Operating under a relaxed but shared assumption that the true target for a given ML task is not assumed to exist as a well-defined object in the real world [11], LAF [13] and UTTL [14] respectively provide the theoretical foundations for the evaluation and learning of predictive models with MIATTs.

### 2.3.1 LAF

LAF provides a logical framework for evaluating predictive models using MIATTs by aggregating multiple partially correct targets [13]. The principle of LAF is: Based on MIATTs, LAF can approximate conventional ATT-based evaluation reasonably well in complex tasks, while potentially exhibiting greater deviations in simpler ones [11, 13]. Thus, LAF can be implemented through logical operations (e.g., conjunction, disjunction, or fuzzy aggregation). It assesses predictive model performance from a multi-perspective logical viewpoint, emphasizing coverage and consistency rather than relying on a single ground truth. This enables fine-grained diagnosis of correctness across different incomplete or uncertain targets.

### 2.3.2 UTTL

UTTL provides a learning paradigm that optimizes predictive models without relying on a single accurate true target [14]. The principle of UTTL is: Based on MIATTs, UTTL can be effectively implemented within a multi-target learning framework [11, 14]. Thus, UTTL can be implemented by learning from multiple incomplete or uncertain targets within the MIATTs set by aligning their shared and reliable components, thereby approximating the underlying true target in a self-consistent and task-adaptive manner.

## 3. Analysis of Task-Specific MIATTs

Regarding the generation and assessment of MIATTs for a specific task, this section analyzes the possible qualities and structural patterns of task-specific MIATTs, along with their downstream influence on evaluation and learning.

### 3.1 Possible qualities of task-specific MIATTs with respect to assessment indicators

Based on the task-specific AIM, multiple diverse MIATTs can be generated according to Formula (2). With respect to the assessment indicators defined in Formulas (5) and (6), the potential qualities of these MIATTs are summarized in Table 1. Specifically, in Table 1, the *mean of PartialRepresentation* reflects the *coverage* of MIATTs with respect to the underlying true target, while *(1 – Redundancy)* represents their *diversity* in capturing the true target.

Table 1. Possible qualities of task-specific MIATTs.

|  | 1-Redundancy=0 | 1-Redundancy=0.5 | 1-Redundancy=1 |
|---|---|---|---|
| mean(PartialRepresentation)=0 | Worst Quality | | |
| mean(PartialRepresentation)=0.5 | | Median Quality | |
| mean(PartialRepresentation)=1 | | | Best Quality |

As shown in Table 1, the quality of task-specific MIATTs is determined jointly by *coverage*—measured by the mean of *PartialRepresentation*—and *diversity*—measured by *(1−Redundancy)*. High coverage ensures that each IATT reliably captures essential aspects of the underlying true target, while high diversity provides complementary perspectives that reduce shared bias. Low values of both indicators lead to the worst quality, where MIATTs fail to represent the underlying true target effectively. Moderate levels yield median quality, offering partial but useful diagnostic information. The best quality is achieved when both coverage and diversity are high, producing MIATTs that are accurate, comprehensive, and mutually reinforcing. This configuration enables robust, fine-grained evaluation and learning, making the MIATTs set a faithful and informative surrogate for the unattainable accurate true target.

## 3.2 Structural patterns of task-specific MIATTs

The structural patterns of task-specific MIATTs corresponding to the quality levels summarized in Table 1 are illustrated in Figure 1. A task-specific MIATTs set can be formed by including all feasible patterns, excluding the unattainable pattern depicted in Figure 1. Each included pattern contributes to capturing different aspects of the underlying true target, with variations in coverage and diversity that collectively determine the overall informativeness and reliability of the MIATTs set. By combining these patterns, the MIATTs set provides a comprehensive, multi-perspective approximation of the underlying true target, enabling robust evaluation and learning.

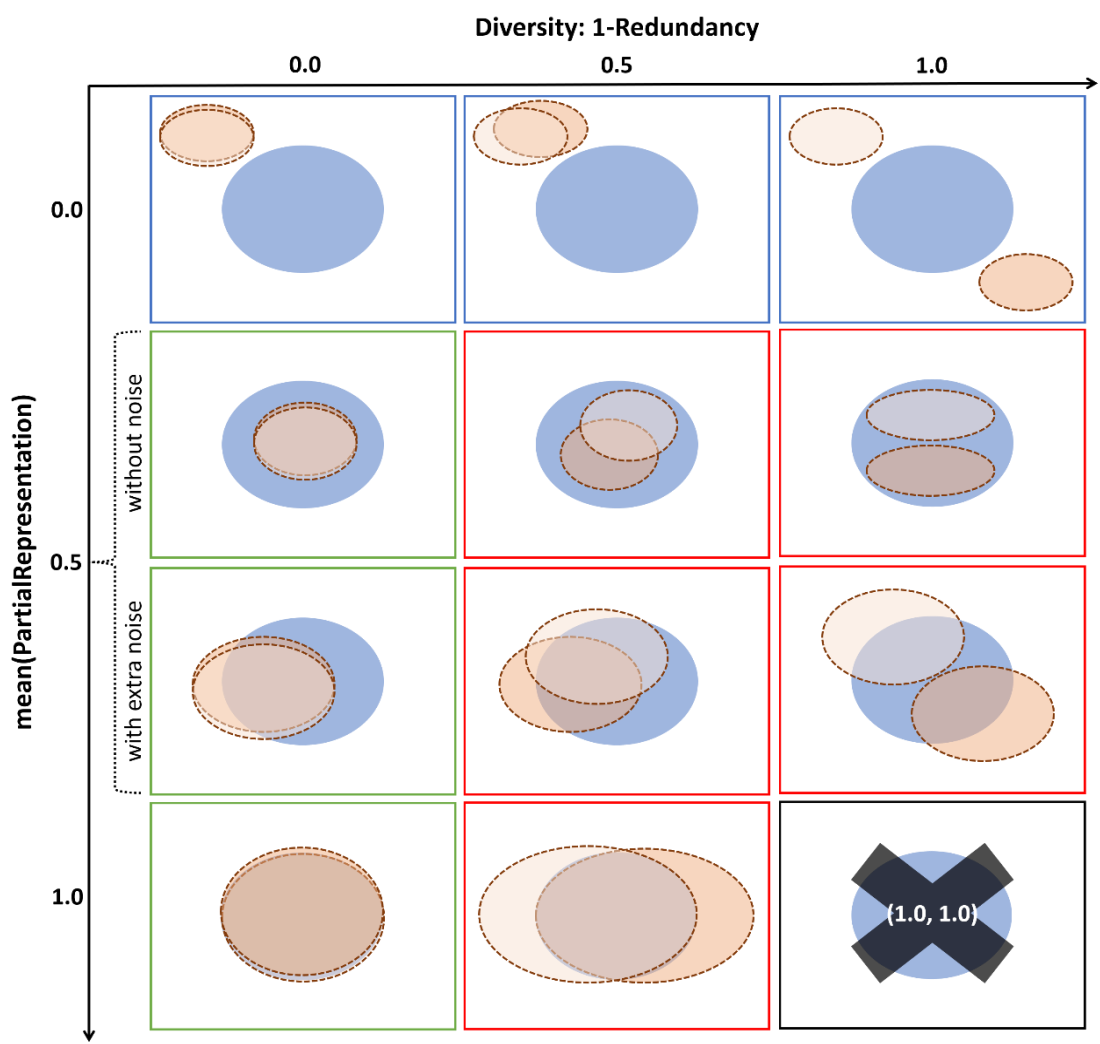

Figure 1. Possible patterns of task-specific MIATTs. Blue circles represent the underlying true target, while orange circles represent the generated MIATTs for approximating it. Blue bounding boxes indicate MIATTs with no coverage. Blue bounding boxes indicate MIATTs with adequate coverage but limited diversity. Red bounding boxes indicate MIATTs with both adequate coverage and diversity. The black bounding boxe indicates the pattern that is unattainable.

### 3.3 Summary of task-specific MIATTs

For a specific task, if the number of MIAs used to generate the task-specific MIATTs is sufficiently large, it is likely that the union of the generated MIATTs set, $\bigcup_{n=1}^{N} t_n^*$, will approximate full coverage of the underlying true target $t^*$. At the same time, this union inevitably introduces some additional noise. Consequently, the resulting MIATTs set provides a near-complete, multi-perspective representation of $t^*$, capturing its essential semantic structure while satisfying the balance between coverage and redundancy.

Thus, the task-specific MIATTs set is subject to
$$SF(t^*) \subseteq \bigcup_{n=1}^{N} SF(t_n^*) \subseteq SF(t^*) \cup \epsilon. \tag{8}$$
And, it can be geometrically visualized as Fig. 2.

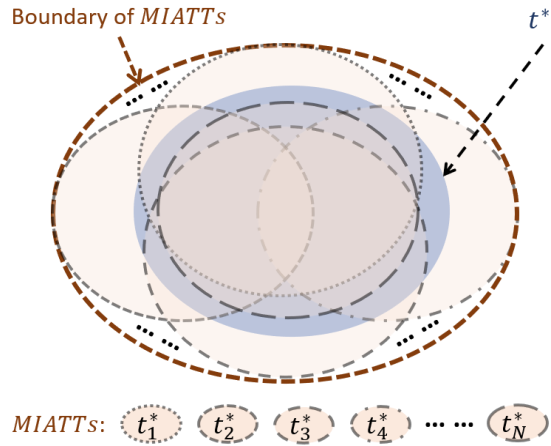

Figure 2. Geometric visualization of task-specific MIATTs.

The relationship depicted by Formula (8) and Fig. 2 reflects a coverage–noise trade-off that propagates into subsequent evaluation and learning processes.

### 3.4 Downstream influence on evaluation and learning

#### 3.4.1 Influence on evaluation (LAF-based)

For evaluation with MIATTs, the coverage completeness of the MIATTs set determines the soundness and stability of LAF aggregation results. High coverage with moderate noise allows the LAF framework—through conjunction, disjunction, or fuzzy t-norm/t-conorm operations—to approximate traditional accurate-target (ATT) evaluation while preserving interpretability. Insufficient coverage leads to incomplete logical representation, increasing the risk of false negatives (correct predictions misjudged as errors). Excessive redundancy or noise, on the other hand, weakens discriminability, potentially smoothing out subtle correctness patterns.

Hence, well-balanced task-specific MIATTs yield evaluation outcomes that are both logically comprehensive and diagnostically informative, aligning with LAF's goal of interpretable uncertainty reasoning.

### 3.4.2 Influence on learning (UTTL-based)

In UTTL-grounded learning, MIATTs act as multiple weak surrogates of the ground truth. When MIATTs exhibit broad but non-redundant coverage, the learning process benefits from multi-target regularization, enhancing robustness and generalization under epistemic uncertainty. Conversely, overlapping or noisy MIATTs may cause inconsistent gradient directions among targets, potentially slowing convergence or leading to bias toward majority patterns. In the extreme case of sparse MIATT coverage, models may underfit due to insufficient supervision signal, emphasizing the need for adaptive weighting or noise correction mechanisms.

Therefore, the statistical behavior of learning with MIATTs depends on the geometric and semantic distribution of the MIATTs set—whether it provides a well-distributed approximation of $t^*$ or an imbalanced, noisy surrogate.

### 3.4.3 Joint implication

The task-specific MIATTs set serves as the epistemic substrate upon which both evaluation and learning are built. Its structural balance between coverage, diversity, and noise governs: the logical fidelity of LAF-based evaluation and the optimization stability of UTTL-based learning. Achieving this balance ensures that EL-MIATTs not only approximate $t^*$ effectively but also sustain coherent interactions between logical semantics and statistical optimization across the full pipeline.

## 4. Evaluation with MIATTs: LAF-Grounded Strategies

Assuming that the true target for a given ML task is not assumed to exist as a well-defined object in the real world [11], LAF provides a principled framework that, based on MIATTs, can approximate ATT-based evaluation effectively in complex settings but may diverge in simpler ones [11, 13]. Accordingly, this section presents LAF-grounded evaluation algorithms employing logical operators such as conjunction, disjunction, and fuzzy aggregation.

Fundamentally, there are two approaches to conducting evaluation with MIATTs: (1) evaluation using the original MIATTs, and (2) evaluation using a ternary target synthesized from MIATTs. The first approach preserves the multi-perspective and partially true nature of each MIATT, enabling detailed diagnostic analysis, whereas the second consolidates these perspectives into a unified, scoreable three-valued target for simplified assessment. In the following section, we systematically discuss and compare these two approaches in terms of their logical foundations, evaluation granularity, interpretability, and computational characteristics.

### 4.1 Evaluation with original MIATTs

Treating semantic facts in MIATTs as computable logical formulas, we use (multi-valued)

logical semantics to evaluate the model's satisfaction with partial true goals and perform reasonable aggregation across multiple incomplete goals. This approach relies on Kleene's three-valued logic (K3) [17–19] and fuzzy logic's t-norm ( ∧ ) and t-conorm ( ∨ ) [15, 16], along with paraconsistency testing [22, 23] and coverage correction [24, 25].

#### 4.1.1 Evaluation metric design (based on logic)

Assume that each inaccurate true target $t_n^*$ is represented by a set of semantic facts (formulas). Each fact $\varphi$ for a single sample $(i, \tilde{t})$ gives a three-valued truth value $v \in \{0, 1/2, 1\}$:

- 1: Satisfied (True);
- 0: Violated (False);
- 1/2: Unknown/Not applicable (Undefined).

1) **Formula level (fact-level):**
- K3 semantics: $\neg a = 1 - a$;
  Conjunction ( ∧ ) uses Gödel t-norm: $a \wedge b = min(a, b)$;
  Disjunction ( ∨ ) using t-conorm: $a \vee b = max(a, b)$;
  Implication: $a \Rightarrow b = max(1 - a, b)$.
- Applicability $A(\varphi)$: Whether the fact is "decidable" for the sample. In practice: $A(\varphi) = 1[v \neq 1/2]$.

2) **Satisfaction of a single IATT $t_n^*$:**
- Intra-fact Aggregation (emphasizing that "partial representations" must be satisfied simultaneously):

$$S_n(i) = \min_{\varphi \in \Phi_n} v_\varphi(i, \tilde{t}). \tag{9}$$

Or a weighted version $S_n(i) = \min_\varphi (\alpha_\varphi \odot v_\varphi)$ (default is equal weighting, $\odot$ is the weighting rule; simpler options include weighted minimum or weighted geometric mean).

- Applicability coverage of this IATT:

$$C_n(i) = \frac{\sum_{\varphi \in \Phi_n} A(\varphi)}{|\Phi_n|} = \frac{\sum_{\varphi \in \Phi_n} 1[v \neq 1/2]}{|\Phi_n|}. \tag{10}$$

3) **Collective coverage of aggregation across multiple MIATTs:**
- We want to reflect "collectively covering more true target facts." Aggregation using t-conorm:

$$S_{MIATTs}(i) = \max_{n=1 \ldots N} S_n(i). \tag{11}$$

Explanation: As long as all the core facts of a model are satisfied, the model is correct in that aspect.

- Alternatively, a probabilistic "at least one aspect is correct" can be used:

$$S_{MIATTs}^{NoisyOR} = 1 - \prod_{n=1}^{N}(1 - S_n(i)). \tag{12}$$

- Overall applicability:

$$C_{MIATTs}(i) = \frac{1}{N}\sum_{n=1}^{N} C_n(i). \tag{13}$$

4) **Paraconsistency penalty:**

When two facts (possibly across MIATTs) with mutually exclusive requirements on the same sample are both "satisfied" to 1 (or close to 1), this is counted as a contradictory hit.
- Define a set of mutually exclusive pairs $M = \{(\varphi_m, \varphi_n)\}$.
- Sample-level contradiction rate:

$$K_{MIATTs}(i) = \frac{1}{|M|} \sum_{(\varphi_m,\varphi_n) \in M} \mathbb{1}[v_{\varphi_m} = 1 \wedge v_{\varphi_n} = 1]. \quad (14)$$

5) **Final sample score (with coverage correction and consistency penalty):**

The final sample score can be expressed as

$$Score_{MIATTs}(i) = \left(\lambda S_{MIATTs}(i) + (1-\lambda)S_{MIATTs}^{NoisyOR}\right) \cdot C_{MIATTs}(i) \cdot \left(1 - \gamma K_{MIATTs}(i)\right), \quad (15)$$

where $\lambda \in [0,1]$ controls "strict correctness in one aspect" vs. "correctness in at least one aspect"; $\gamma \in [0,1]$ controls the intensity of the contradiction penalty.

6) **Dataset-level metrics:**
- Average:

$$\overline{Score}_{MIATTs} = \frac{1}{|D|} \sum_{i \in D} Score_{MIATTs}(i). \quad (16)$$

- Simultaneously output $\bar{C}_{MIATTs} = \frac{1}{|D|} \sum_{i \in D} C_{MIATTs}(i)$ ("decidability" of the evaluation), $\bar{K}_{MIATTs} = \frac{1}{|D|} \sum_{i \in D} K_{MIATTs}(i)$ (overall contradiction rate).

### 4.1.2 Methodological key points and scalability

This logic-based metric design method for evaluation with original MIATTs has following key points and scalability.

1) **Aligned with the definition:**
   - "Partial representation" → Use conjunction (min) within a single IATT to force the key facts in that aspect to hold simultaneously.
   - "Collective coverage" → Use disjunction (max / NoisyOR) between MIATTs to indicate that "together they cover more aspects of the truth."
2) **Unknown/Not applicable:**
   Use $Undefined = 0.5$ to maintain the good algebraic properties of K3; the coverage $C$ allows you to determine "how many facts were used" in the evaluation.
3) **Conflict resolution:**
   Mutually Exclusive Group + Penalty Term $(1 - \gamma K_{MIATTs})$ is a simple "quasi-parallel consistency" approach; complex scenarios can be replaced with Belnap four-valued logic ($\{\bot, \top, both, neither\}$) or a constraint solver [26, 27].
4) **Weight learning:**
   Weights can be given by an expert or meta-learned using a validation set (e.g., using the $\overline{Score}_{MIATTs}$ as the target and using Bayesian Optimization/Differential Evolution to find weights).
5) **Interpretability:**
   Each sample has a corresponding set of multiple true targets (MIATTs) for evaluation, which naturally provides an explanation of the hotspots of "where it satisfies/does not

satisfy."

## 4.2 Evaluation with ternary target synthesized from MIATTs

We combine MIATTs into a three-valued logical "synthetic true target" $t^†$ via logical merging, then we use this $t^†$ to evaluate the quality of the model's predictions.

### 4.2.1 Logical merging idea

MIATTs are characterized by the following characteristics: each MIATT $t_n^*$ captures only part of the facts of the true target. When merging, we hope to form a single three-valued logical target $t^†$, which can express:
- *True* (1): All MIATTs agree that this is true for this sample;
- *False* (0): All MIATTs agree that this is false for this sample;
- *Unknown* (1/2): When MIATTs have incomplete or conflicting information.

Then the formalization of this idea can be: Suppose that each MIATT $t_n^*$ is given a true value $v_n \in \{0, 0.5, 1\}$ on the sample $(i, \tilde{t})$. The true value of the synthetic target $t^†$ is:

$$t^†(i, \tilde{t}) = \begin{cases} 1, & if\ \forall n, v_n = 1 \\ 0, & if\ \forall n, v_n = 0 \\ 0.5, & otherwise\ (mixed/unknown) \end{cases}. \tag{17}$$

Thus $t^†$ becomes a three-valued logic true target and can be used directly to evaluate the model.

### 4.2.2 Evaluation algorithm based on three-valued logic true targets

1) **Fact computation**
- For each sample $(i, \tilde{t})$, calculate the true value of all MIATTs $\{v_n\}$;
- Use the above rules to generate the composite truth value $t^†(i, \tilde{t})$.

2) **Model score definition**
- For a sample, the score is:

$$Score_{t^†}(i) = t^†(i, \tilde{t}). \tag{18}$$

That is:
- ✧ If target = 1, the prediction is completely correct → score 1;
- ✧ If target = 0, the prediction is completely wrong → score 0;
- ✧ If target = ½, the prediction is "uncertain/insufficient" given the partial true target → score 0.5.
- Overall score for the dataset:

$$\overline{Score}_{t^†} = \frac{1}{|D|} \sum_{i \in D} Score_{t^†}(i). \tag{19}$$

## 4.3 Comparison between the two methods for evaluation with MIATTs

Evaluation with the original MIATTs relies on a parallel multi-perspective logic, emphasizing coverage and multifaceted correctness, and enabling fine-grained diagnostic analysis. In contrast, evaluation with a ternary target synthesized from MIATTs is grounded in a synthetic three-valued logic goal, yielding a compressed single truth target that trades detailed information for a unified, directly scoreable measure. The former resembles logical set semantics, where multiple partial truths coexist in parallel, while the latter approximates a ground truth within a multi-valued logic framework. A detailed comparison of the two approaches to MIATTs-based evaluation is provided in Table 2.

Table 2. Comparison of two approaches to MIATTs-based evaluation

| Feature | Evaluation with original MIATTs | Evaluation with ternary target synthesized from MIATTs |
|---|---|---|
| Granularity | Preserves the "partial truth" of each MIATT, showing where the model performs well/poorly | Merged into a single unified target, losing detailed aspects |
| Truth space | Multiple values ({0, 0.5, 1}) + aggregation logic (min/max/Noisy-OR) | Single value ({0, 0.5, 1}) |
| Information loss | **Low**; allows diagnosing "on which MIATT the model fails" | **High**; conflicts/partial coverage compressed into "0.5" |
| Interpretability | **Strong**; directly explains "on which inaccurate target the model does well/poorly" | **Weak**; only states "correct/incorrect/uncertain" |
| Computational simplicity | **Complex**; requires handling coverage, contradiction rate, etc. | **Simple**; only needs merging rules |
| Use case | Suitable for analysis/research scenarios (high interpretability) | Suitable for scenarios requiring a single metric (e.g., quick model comparison) |

## 4.4 Relation between MIATTs-based and ATT-based evaluation

We examine the logical and mathematical relations between the two MIATTs-based evaluation approaches and the standard accurate true target (ATT)-based evaluation.

### 4.4.1 Baseline: ATT $t^*$

Assuming the underlying true target $t^*$ is fully definable, the evaluation is the same as normal evaluation with ATT:

$$Score_{t^*}(i) = \begin{cases} 1 & \tilde{t} = t^*(i) \\ 0 & \tilde{t} \neq t^*(i) \end{cases}. \tag{20}$$

Or in the continuous value scenario, use $|\tilde{t} - t^*(i)|$, log-likelihood, etc.

Logically, this is equivalent to having a complete Boolean truth function, without ambiguity or unknowns.

### 4.4.2 Relation between original MIATTs and baseline

Logical Relationship: The whole set of original MIATTs are partial projections of $t^*$ and with additional noise ($SF(t^*) \subseteq \bigcup_{n=1}^{N} SF(t_n^*) \subseteq SF(t^*) \cup \epsilon$). Evaluation is equivalent to checking whether the model satisfies several logical clauses. The aggregation method (min/max/NoisyOR) determines whether we approximate $t^*$ more conservatively or more loosely.

Mathematically: The whole set of original MIATTs is an upper approximation of $t^*$ with additional noise ($SF(t^*) \subseteq \bigcup_{n=1}^{N} SF(t_n^*) \subseteq SF(t^*) \cup \epsilon$). If the model is correct or error on all MIATTs, it must be correct or error on $t^*$ (a sufficient condition). But if the model is correct or error on a particular IATT, it does not necessarily mean it is correct or error on $t^*$ (it may simply have missed another aspect).

Thus, evaluation with original MIATTs is complete but unsound: while it preserves the full information of $t^*$, it can sometimes misclassify correct results as errors and errors as correct.

### 4.4.3 Relation between ternary target synthesized from MIATTs ($t^\dagger$) and baseline

Logical Relationship: The ternary target synthesized from MIATTs ($t^\dagger$) combines the results of multiple partial targets $t_n^*$ into a single three-valued domain:
- 1: All aspects are consistently true → consistent with $t^*$;
- 0: All aspects are consistently false → consistent with $t^*$;
- 0.5: Inconsistent or insufficient information → indicates that we cannot determine the relationship with $t^*$.

Mathematically: The $t^\dagger$ is equivalent to introducing an "uncertain" intermediate value on the classical truth value set {0,1}. Thus, it is equivalent to constructing a three-valued upper approximation of $t^*$ with uncertainty:
- If $t^\dagger = 1$, then $t^* = 1$ (reliable);
- If $t^\dagger = 0$, then $t^* = 0$ (reliable);
- If $t^\dagger = 0.5$, then $t^*$ may be 0 or 1 (uncertain).

So $t^\dagger$ is partial but consistent: it will not mistake an error for a correct answer, nor a correct answer for an error, but it may abandon the judgment.

### 4.4.4 Relation between the three

Intuitive analogy:
- $t^*$: Given a complete map, determine whether the path is correct.
- MIATTS: Given only a few partial maps with noise, you can say "right/wrong" on these partial paths, but you can't guarantee the overall picture.
- $t^\dagger$: Overlay all partial maps to create a simplified map; any conflicts or omissions are marked as "unknown."

More detailed relation between the three are provided in Table 3.

Table 3. Summarized relation between the three methods for evaluation.

| Method | Mathematical relation to $t^*$ | Property | Misjudgment risk | Information loss |
|---|---|---|---|---|
| $t^*$ | Exact equivalence | Precise ground truth | None | None |
| MIATTs | Upper approximation (with extra noisy zone) | Complete but unsound | May sometimes yield misclassify correct results as errors and errors as correct | Very high; probably can preserve the entire information of $t^*$ |
| Synthesized ternary $t^\dagger$ | Upper approximation (with uncertainty zone) | Partial but consistent | No misjudgment, but produces "uncertain" cases | High; some details compressed into uncertain (0.5) |

The geometry relation between the three can be shown as Fig. 3.

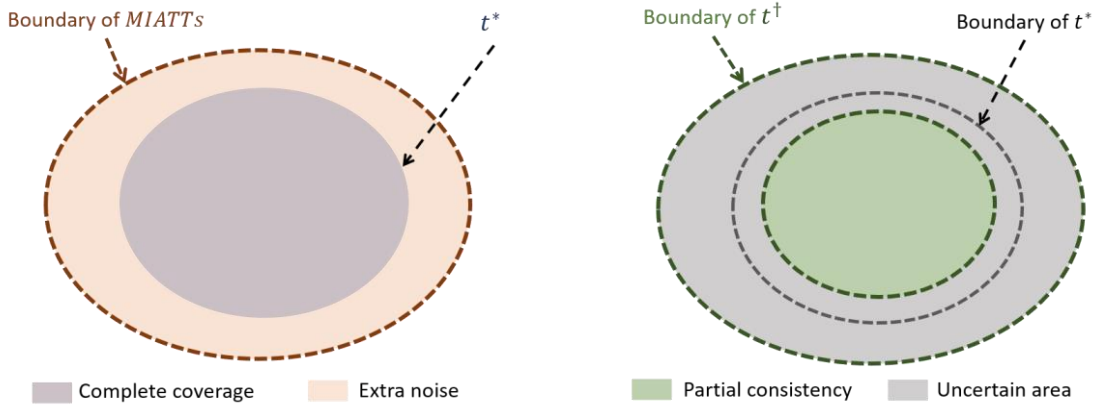

Figure 3. Geometry relation between the three. Left: Upper approximation (with extra noisy zone) of MIATTs to $t^*$. Right: Upper approximation (with uncertain zone) of synthesized ternary $t^\dagger$ to $t^*$.

#### 4.4.5 Summary

Direct evaluation with ATT ($t^*$) is the gold standard (but often unavailable). Evaluation with original MIATTs sacrifices unsoundness for completeness and interpretability under "multiple partial knowledge" with additional noise. Evaluation with triple-valued synthetic target ($t^\dagger$) sacrifices detail for a single, unified, three-valued judgment. Therefore, evaluation with original MIATTs is more suitable for diagnostic analysis (which aspects are correct/incorrect). Triple-valued synthetic method is better for quickly comparing models overall, but they discard judgments on some samples. Both can be viewed as approximate logical alternatives to true evaluation when the $t^*$ is missing.

## 5. Learning with MIATTs: UTTL-Grounded Strategies

Assuming that the true target for a given ML task is not assumed to exist as a well-defined object in the real world [11], UTTL provides a principled framework for learning from MIATTs, which indicates that UTTL can be effectively implemented through a multi-target learning paradigm [11, 14]. Accordingly, this section introduces UTTL-grounded learning strategies constructed upon two commonly used loss functions—Dice [20] and Cross Entropy (CE) [21].

When handling multiple targets (e.g., multiple annotators or probabilistic labels), two general aggregation strategies can be adopted: Method A (Per-target then Aggregate): compute the loss for each target separately and then aggregate the individual losses (e.g., through averaging or robust weighting); and Method B (Aggregate then Single Loss): first aggregate all targets (e.g., by averaging) and then compute a single loss with respect to the aggregated target. Although these two approaches may appear similar in form, their theoretical behavior diverges depending on the specific loss function. In the following discussion, we analyze and summarize the characteristics and implications of both methods under Dice and CE loss functions.

## 5.1 Comparison of Dice-based loss formulations for Methods A and B

Let $\tilde{t} \in (0,1)$ denote the model prediction for a given pixel (or soft mask value), and let $MIATTs = \{t_n^* | n \in \{1, \cdots, N\}\} \in \{0,1\}$ denote for multiple inaccurate true targets (e.g. multiple soft targets or multiple measurement repeats). Define the per-pair Dice similarity

$$Dice(\tilde{t}, t^*) = \frac{2\tilde{t}t^*}{\tilde{t}^2 + t^{*2}}. \tag{21}$$

We compare the following two training objectives (losses rewritten in "score" form $Dice$; if you use $1 - Dice$ as the loss the inequalities below reverse accordingly):

- **Averaged per-target Dice** (Avg-Dice, Method A):

$$S_A(\tilde{t}, MIATTs = \{t_n^* | n \in \{1, \cdots, N\}\}) = \frac{1}{N}\sum_{n=1}^{N} Dice(\tilde{t}, t_n^*). \tag{22}$$

- **Dice of averaged targets** (Dice-of-Mean, Method B):

$$S_B(\tilde{t}, MIATTs = \{t_n^* | n \in \{1, \cdots, N\}\}) = Dice(\tilde{t}, \overline{t^*}), \quad \overline{t^*} = \frac{1}{N}\sum_{n=1}^{N} t_n^*.$$

Below we analyze their mathematical relationships, optimization implications, robustness and practical recommendations.

### 5.1.1 Mathematical relation (Jensen bias) between Methods A and B

Treating $\tilde{t}$ as fixed, consider $Dice$ as a function of $t^*$. Its second derivative (with respect to $t^*$) is

$$\frac{\partial^2 Dice}{\partial^2 t^{*2}} = \frac{4\tilde{t}t^*(t^{*2} - 3\tilde{t}^2)}{(\tilde{t}^2 + t^{*2})^3}. \tag{23}$$

Consequently $Dice$ is concave in $t^*$ for $t^{*2} < 3\tilde{t}^2$ (i.e. $t^* < \sqrt{3}\tilde{t}$) and convex for $t^{*2} > 3\tilde{t}^2$. By Jensen's inequality:

- If all $t_n^*$ lie in a concave region for the give $\tilde{t}$ then

$$S_B = Dice(\tilde{t}, \overline{t^*}) \geq \frac{1}{N}\sum_{n=1}^{N} \emptyset(\tilde{t}, t_n^*) = S_A. \tag{24}$$

Equivalently, using losses $1 - Dice$, Dice-of-Mean underestimates the average loss (is overly optimistic).

- If all $t_n^*$ lie in a convex region, the inequality reverses:

$$S_B \leq S_A, \tag{25}$$

so Dice-of-Mean is more conservative.

- If the $t_n^*$ span both regions (some below and some above $\sqrt{3}\tilde{t}$), no universal ordering holds; numerical comparison is required.

Equality $S_B = S_A$ holds if $t_1^* = t_2^* = \cdots = t_N^*$ (or $Dice$ is affine over the convex hull of $t_n^*$, which here only occurs at a degenerate set).

Dice-of-Mean (Method B) introduces a systematic bias relative to the mean per-target Dice (Method A) because $Dice$ is nonlinear in $t^*$; the direction of bias depends on where the $t_n^*$ lie relative to the threshold $\sqrt{3}\tilde{t}$.

### 5.1.2 Gradient and optimization dynamics

Compute derivative of $Dice$ with respect to $\tilde{t}$:

$$\frac{\partial Dice(\tilde{t}, t^*)}{\partial \tilde{t}} = \frac{2t^*(t^{*2} - \tilde{t}^2)}{(\tilde{t}^2 + t^{*2})^2}. \tag{26}$$

Thus, gradients under the two schemes are

- Avg-Dice (Method A):

$$\nabla_{\tilde{t}} S_A = \frac{1}{N} \sum_{n=1}^{N} \frac{2t_n^*\left(t_n^{*2}-\tilde{t}^2\right)}{\left(\tilde{t}^2+t_n^{*2}\right)^2}. \tag{27}$$

- Dice-of-Mean (Method B):

$$\nabla_{\tilde{t}} S_B = \frac{2\bar{t}^*\left(\bar{t}^{*2}-\tilde{t}^2\right)}{\left(\tilde{t}^2+\bar{t}^{*2}\right)^2}. \tag{28}$$

Consequences: $S_A$ aggregates per-target gradients and therefore preserves heterogeneity: extremes in individual $t_n^*$ produce strong corrective signals for $\tilde{t}$. This supports faster correction of individual mismatches and is beneficial when multiple targets are informative but heterogeneous. $S_B$ uses a single gradient computed at $\bar{t}^*$. Strong individual signals from outlying $t_n^*$ are diluted by averaging; in effect, Dice-of-Mean yields lower gradient variance but may slow correction of biased/erroneous targets or rare but important modes.

### 5.1.3 Robustness to target noise and outliers

Avg-Dice (Method A) is sensitive to outliers: a single erroneous $t_n^*$ contributes directly to the average and can dominate the update if its gradient magnitude is large. However, this sensitivity is controllable, one can apply per-target weights, trimmed means, or robust aggregators to mitigate outliers while retaining per-target supervision.

Dice-of-Mean (Method B) naturally smooths isolated target noise via the averaging step, so single outliers have reduced immediate impact. Nevertheless, because of the nonlinearity of $Dice$, a sufficiently large outlier may still shift $\bar{t}^*$ across the convexity threshold and induce a nontrivial bias.

Recommendation: If annotator reliability is unknown and outliers are expected, use $S_A$ with robust aggregation or weighted averaging; use $S_B$ only when target noise is approximately zero-mean and targets are exchangeable.

### 5.1.4 Interpretability and diagnostic capability

$S_A$ provides per-target loss terms that are directly interpretable and diagnostic: one can detect which targets systematically disagree with the model and apply target-specific strategies (reweighting, calibration, or retraining). $S_B$ obviates per-target diagnostics because only the aggregate label is used; this simplifies implementation but loses visibility into target heterogeneity

### 5.1.5 Computational considerations

Computational cost difference is minor in typical deep learning settings (vectorized implementations make multiple evaluations cheap). $S_B$ does one Dice computation; $S_A$ does $N$. If the number of targets is very large, cost matters and one may prefer an aggregated target. Otherwise, cost should not drive the choice.

### 5.1.6 Practical recommendations

(1) When to prefer $S_A$ (Avg-Dice)
    o Multiple targets with unknown or varying reliability.
    o Need for per-target diagnostics, reweighting or curriculum learning.
    o Desire to preserve strong corrective signals from individual labels (e.g. rare features or small structures).

(2) When to prefer $S_B$ (Dice-of-Mean)
    o Labels are many noisy observations drawn from a common unbiased process

and computational simplicity is desired.
- The training objective must match an inference scheme that uses averaged labels/predictions (i.e. you care about optimizing the mean output directly).

(3) Hybrid and robust schemes
- Use $S_A$ together with a robust aggregator on the loss (e.g. trimmed mean, median of means).
- Aggregate labels with a reliability-weighted mean $\bar{t}^* = \sum_{n=1}^{N} w_n t_n^*$ (learn or estimate $w_n$), then apply $Dice(\tilde{t}, \bar{t}^*)$. This can combine the smoothing of $S_B$ with weighted robustness.
- Combine both terms in a composite loss:

$$L = \lambda(1 - S_A) + (1 - \lambda)(1 - S_B), \tag{29}$$

where $\lambda \in [0,1]$ trades per-target supervision against ensemble-level optimization.

(4) **Validation monitoring**
- Monitor per-target metrics and ensemble metrics on a held-out set to detect Jensen bias and divergence between $S_A$ and $S_B$ behaviour.

### 5.1.7 Concise takeaway

Because Dice similarity is nonlinear in targets, averaging before applying Dice (Dice-of-Mean, Method B) is not algebraically equivalent to averaging per-target Dice (Avg-Dice, Method A). The two choices encode a bias–variance trade-off:

✓ $S_A$ preserves target-specific corrective signals (higher variance, greater fidelity), whereas $S_B$ reduces variance via averaging but can introduce a systematic bias whose sign depends on the local concavity/convexity of $Dice$ relative to $\tilde{t}$.

✓ In most practical annotation settings—where targets differ and diagnostics are desirable—$S_A$ (possibly with robust aggregation or weighting) is the safer default; $S_B$ may be chosen when targets are exchangeable, noise is symmetric, and simplicity or inference-matching is prioritized.

## 5.2 Comparison of CE-based loss formulations for Methods A and B

Let's focus our discussion on binary/multiclass cross-entropy (CE) and rigorously compare two approaches:

- **Averaged per-target CE** (Avg-CE, Method A):

$$S_A\big(\tilde{t}, MIATTs = \{t_n^* | n \in \{1, \cdots, N\}\}\big) = \frac{1}{N} \sum_{n=1}^{N} CE(\tilde{t}, t_n^*). \tag{30}$$

- **Dice of averaged targets** (CE-of-Mean, Method B):

$$S_B\big(\tilde{t}, MIATTs = \{t_n^* | n \in \{1, \cdots, N\}\}\big) = CE(\tilde{t}, \bar{t}^*), \ \ \bar{t}^* = \frac{1}{N} \sum_{n=1}^{N} t_n^*. \tag{31}$$

Below are the conclusions, algebraic proofs, and practical considerations.

### 5.2.1 Conclusions

For standard cross entropy (no additional nonlinear transformations, no weighting terms, and targets with probability distributions or soft labels), Method A and Method B are completely equivalent: their numerical values, gradients, and optimization impact are identical. Therefore, in this case, there is no Jensen bias issue, as with Dice.

However, in terms of implementation and engineering, Method A still retains more diagnostic/weighting flexibility (for example, the ability to weight or prune individual targets), while Method B is more concise but tends to "obscure" individual information.

### 5.2.2 Algebraic proof (Taking binary CE as an example)

Define binary CE:
$$CE(\tilde{t}, t^*) = -[t^* \log \tilde{t} + (1 - t^*) \log(1 - \tilde{t})], \tag{32}$$

where $\tilde{t} \in (0,1)$ is the model output probability, and $t^* \in [0,1]$ is the target probability (which can be a soft target or target average). Expand $CE(\tilde{t}, t^*)$ by $t^*$:
$$CE(\tilde{t}, t^*) = -t^* \log \tilde{t} - (1 - t^*) \log(1 - \tilde{t}) = t^*(-\log \tilde{t} + \log(1 - \tilde{t})) - \log(1 - \tilde{t}).$$

It can be seen that $CE(\tilde{t}, t^*)$ is an affine (linear) function with respect to $t^*$ of the form $a(\tilde{t})t^* + b(\tilde{t})$.

Thus, for a set of $t_n^*$:
$$CE(\tilde{t}, \overline{t^*}) = a(\tilde{t})\overline{t^*} + b(\tilde{t}) = a(\tilde{t})\frac{1}{N}\sum_{n=1}^{N} t_n^* + b(\tilde{t}) = \frac{1}{N}\sum_{n=1}^{N}\left(a(\tilde{t})t_n^* + b(\tilde{t})\right) =$$

$$\frac{1}{N}\sum_{n=1}^{N} CE(\tilde{t}, t_n^*). \tag{33}$$

That is
$$S_A = S_B. \tag{34}$$

### 5.2.3 Gradient level (Proving consistency)

Taking the derivative with respect to $\tilde{t}$, the derivative of the binary CE is linear with respect to $t^*$:
$$\frac{\partial CE(\tilde{t}, t^*)}{\partial \tilde{t}} = \frac{\tilde{t} - t^*}{\tilde{t}(1 - \tilde{t})}. \tag{35}$$

Therefore,
$$\nabla_{\tilde{t}} S_A = \frac{1}{N}\sum_{n=1}^{N} \frac{\tilde{t} - t_n^*}{\tilde{t}(1 - \tilde{t})} = \frac{\tilde{t} - \overline{t^*}}{\tilde{t}(1 - \tilde{t})} = \nabla_{\tilde{t}} S_B. \tag{36}$$

So the gradients are exactly the same, and the training behavior is equivalent.

### 5.2.4 Multi-class case

If we use multi-class cross entropy (categorical CE) and consider as class probability vectors, the same holds true: CE is linear in the target distribution, so
$$CE\left(\tilde{t}, \frac{1}{N}\sum_{n=1}^{N} t_n^*\right) = \frac{1}{N}\sum_{n=1}^{N} CE(\tilde{t}, t_n^*). \tag{37}$$

### 5.2.5 Practical notes and exceptions

Although mathematically equivalent, engineering/practical considerations require the following:

(1) Additive weights or nonlinear postprocessing can violate the equivalence: If different weights $w_n$ are introduced to each target's loss, then $CE(\tilde{t}, \overline{t^*})$ is no longer equal to $\sum_{n=1}^{N} w_n CE(\tilde{t}, t_n^*)$ (unless the weights are equal and $\overline{t^*}$ are weighted equally). If nonlinear transformations (such as truncation, thresholding, temperature scaling, or logit-space operations) are performed on $t_n^*$ before computing the loss, the equivalence disappears.

(2) Robust aggregation and outlier handling: If robust aggregations on outliers, such as de-extinction, mean truncating, or median truncation are performed, aggregating first and then performing CE will produce different results than performing CE first and then performing robust operations (because robust operations are often nonlinear). While mathematically equivalent, from an engineering perspective, you may want to retain the loss of a single target for diagnostic purposes, which is more convenient with Method A (although computationally, vectorization can recover the same results).
(3) Target calibration/temperature/target smoothing: If target smoothing or temperature scaling is performed on the averaged soft targets, the behavior will differ from performing CE first and then averaging on each $t_n^*$ (this can differ depending on the order).
(4) Precision and numerical stability: Due to floating-point arithmetic, the numerical errors between the two implementations may differ very slightly at extreme scales, but these are usually negligible.

### 5.2.6 Concise takeaway

Advices for usage of the CE-based Methods A and B can be summarized as follows:
- ✓ If the standard (unweighted, no additional nonlinearity) cross-entropy is used, the two methods are theoretically and optimizationally equivalent, so Methods A or B can be chosen based on implementation preference (averaging the labels first and then computing the loss is convenient for saving memory or simplifying code).
- ✓ If diagnosing target behavior, weighting individual targets, or providing robustness to individual losses are needed, method A (single loss first, then aggregation) is more flexible—even though the mathematical equivalent is true, method A is easier to scale and debug.
- ✓ If post-processing of the target (thresholding, truncation, temperature, weighting, etc.) is included in pipeline, be sure to re-analyze which order makes sense, as this will break the equivalence.

## 5.3 Summarization of Methods A and B under Dice and CE

For Dice loss, Method A should be regarded as the default due to its preservation of target-specific corrective signals and its compatibility with weighting or robust aggregation schemes. While Method B can reduce variance and may simplify implementation, it inevitably introduces systematic bias, making it less reliable in heterogeneous annotation settings For CE loss, the situation differs fundamentally: under the standard form (no weighting, no nonlinear label transformations), Method A and Method B are mathematically and optimizationally equivalent. Thus, practical considerations such as memory footprint, code simplicity, or diagnostic flexibility should guide the choice. Importantly, once additional operations (e.g., temperature scaling, thresholding, or per-target weighting) are applied, the equivalence is broken, and Method A again becomes the safer and more generalizable choice. See Table 4.

Table 4. Summarization of Methods A and B under Dice and CE loss

| Loss | Method A | Method B | Equivalence & Bias | Recommended Usage |
|---|---|---|---|---|
| **Dice** | Computes Dice per target and aggregates thereafter; preserves target-specific corrective signals, maintaining fidelity at the cost of higher variance. | Aggregates targets first, then computes a single Dice score; variance is reduced but bias is introduced. | **Not equivalent.** Due to nonlinearity of Dice, Method B incurs Jensen-type bias. The bias direction depends on local concavity/convexity of Dice with respect to the averaged target. | Method A is safer in most practical annotation settings, especially when target-specific diagnostics or robustness are needed. Method B may be acceptable when targets are exchangeable, annotation noise is symmetric, and implementation simplicity is prioritized. |
| **CE (standard)** | Computes CE per target (with probability distributions or soft labels), then aggregates. Provides hooks for weighting, pruning, or diagnostics. | Aggregates targets/labels first, then computes a single CE. Compact and memory-efficient. | **Completely equivalent** under standard CE: identical values, gradients, and optimization behavior. No Jensen bias issue. | Either method is acceptable for standard CE; choice can follow implementation preference. Method A remains advantageous when post-hoc weighting, target-specific robustness, or diagnostic analysis are required. Equivalence breaks if nonlinear post-processing (e.g., thresholding, temperature scaling, weighting) is applied to the targets. |

## 6. Discussion

The concept of MIATTs provides a pragmatic response to the epistemological limitation that the true target of a ML task is not assumed to exist as a well-defined object in the real world. Instead, MIATTs represent diverse, partially correct, and complementary views derived from task-specific AI models (AIM). The quality of a task-specific MIATTs set can be characterized by two key indicators: *mean(PartialRepresentation)*, representing its coverage of the underlying true target, and *(1-Redundancy)*, reflecting its diversity. As summarized in Table 1, higher coverage and lower redundancy together yield higher-quality MIATTs, while limited diversity or insufficient coverage leads to degraded representational fidelity.

When the number of MIAs involved in generating MIATTs increases, the union of all MIATTs ($\cup_{n=1}^{N} t_n^*$) tends to approximate full coverage of the latent true target $t^*$, though this inevitably introduces noise. Hence, an optimal MIATTs configuration should achieve a balance between completeness and consistency—maximizing semantic coverage while minimizing overlap and contradiction. This balance ensures that MIATTs collectively provide a robust, multi-perspective representation suitable for both evaluation and learning.

The downstream influence of task-specific MIATTs is twofold. For evaluation, the balance between coverage and redundancy directly affects the soundness–completeness trade-off: higher coverage enhances completeness but may reduce soundness due to noise, while moderate redundancy supports more stable, interpretable scoring under LAF-based evaluation. For learning, MIATTs diversity provides richer supervision signals that improve generalization and robustness in UTTL-grounded strategies, though excessive overlap or noise can introduce gradient conflicts or bias. Hence, maintaining a balanced MIATTs structure—adequate coverage with controlled redundancy—is essential for stable and reliable model evaluation and training within the EL-MIATTs framework.

From the evaluation perspective, LAF offers a principled foundation for assessing models when ground truth cannot be assumed as a single precise label. Under LAF, evaluation can proceed in two main forms: (1) Evaluation using original MIATTs, which retains each MIATT's partial truth and employs logical or fuzzy operators (e.g., conjunction, disjunction, t-norms, and t-conorms) for aggregation; and (2) Evaluation using a ternary target synthesized from MIATTs, which compresses multi-perspective truth into a unified three-valued label {0,0.5,1}. The former provides fine-grained interpretability and preserves detailed coverage information, enabling diagnosis of where and why a model performs well or poorly. However, it is computationally complex and may yield inconsistent judgments across MIATTs. The latter approach simplifies computation and facilitates single-metric comparison but sacrifices information granularity.

When using original MIATTs, LAF-grounded evaluation can be viewed as complete but not perfectly sound—it captures the full informational spectrum of $t^*$ yet may misinterpret partially correct predictions as errors or vice versa. Nevertheless, when combined with appropriate fuzzy aggregation (e.g., using min/max or Noisy-OR operators), LAF enables a balanced trade-off between interpretability, consistency, and practicality. In contrast, when using a ternary target synthesized from MIATTs, evaluation becomes sound but incomplete—the unified three-valued representation ensures consistency and ease of comparison but inevitably compresses detailed multi-perspective information into a simplified truth structure.

This trade-off enhances computational efficiency and interpretability at the global level but reduces diagnostic depth regarding which MIATT contributed to a specific outcome. Overall, LAF-grounded evaluation—whether using original MIATTs or a synthesized ternary target—offers complementary perspectives. In complex, ill-defined tasks, the former better approximates traditional accurate-target evaluation by capturing nuanced partial truths, whereas in simpler or more deterministic tasks, the latter provides a more stable and pragmatic alternative. Deviations observed in either case largely reflect the intrinsic ambiguity of the task itself rather than methodological inadequacy.

On the learning side, UTTL establishes a theoretical foundation for training models when the true target cannot be precisely defined. Within UTTL, MIATTs are treated as multiple weakly reliable surrogates of the underlying truth, allowing learning to proceed through multi-target optimization rather than single-target supervision. Two general strategies can be adopted: Method A (Per-target then Aggregate) computes the loss with respect to each MIATT separately and aggregates the individual losses (e.g., via averaging or weighted combination). Method B (Aggregate then Single Loss) first aggregates all MIATTs into a composite target and then computes a single loss against it.

When implemented with Dice and Cross Entropy (CE) loss functions, these strategies exhibit distinct theoretical behaviors. Method A emphasizes robustness and diversity by allowing individual MIATT contributions to remain explicit, making it suitable when MIATTs vary substantially in reliability. Method B, by contrast, enforces global consistency and is computationally simpler but risks losing fine-grained information. UTTL therefore provides a flexible unifying framework in which both strategies can be adapted to task complexity, data uncertainty, and desired learning bias.

Together, LAF-based evaluation and UTTL-based learning bridge the theoretical gap between logical semantics and statistical optimization in machine learning under uncertain supervision. The two are mutually reinforcing components within the EL-MIATTs paradigm. LAF provides a logic-grounded framework for evaluating model outputs against multiple partially true targets, ensuring that model assessment remains consistent with multi-valued truth semantics rather than collapsing ambiguity into binary correctness. UTTL, in turn, translates these same logical principles into a statistical learning mechanism, optimizing model parameters in a way that reflects the graded or composite nature of truth encoded by MIATTs.

Through this integration, logical reasoning—expressed via conjunction, disjunction, and fuzzy aggregation in LAF—is aligned with the probabilistic gradient-based optimization used in UTTL. This alignment allows evaluation and learning to operate within a shared semantic space, where logical consistency and statistical efficiency jointly contribute to model improvement. Consequently, EL-MIATTs not only evaluate but also train models in a manner that respects uncertainty, disagreement, and partial correctness, offering a unified pathway toward explainable, epistemically grounded, and practically effective machine learning.

## 7. Conclusion, Limitation, and Future Work

This paper presents a systematic effort to bridge the theoretical formulation and practical implementation of the EL-MIATTs framework [11], emphasizing how LAF-based evaluation algorithms and UTTL-based learning strategies enable reliable model assessment and training under epistemic uncertainty. By grounding the evaluation process in the LAF and the learning process in the UTTL paradigm [13, 14], we operationalize EL-MIATTs into two coherent, complementary mechanisms.

Specifically, we analyzed the structural and semantic properties of task-specific MIATTs [12] and its downstream influence on evaluation and learning, proposed two LAF-grounded evaluation schemes (parallel multi-perspective and ternary synthesized), and developed two UTTL-grounded learning strategies (Per-target then Aggregate and Aggregate then Single Loss) applicable to common loss functions such as Dice and Cross Entropy. Together, these developments enhance the robustness, interpretability, and adaptability of predictive models in settings where the "true" target is uncertain, incomplete, or inherently multi-faceted. Furthermore, the discussion about the integration of LAF and UTTL illustrates how logical semantics and statistical optimization can be unified within the EL-MIATTs framework, offering a principled pathway toward more explainable and uncertainty-aware machine learning.

Despite its theoretical soundness and practical flexibility, the current implementation of EL-MIATTs remains subject to several limitations. First, the construction of task-specific MIATTs depends on the availability and diversity of task-relevant AI models (AIM), which may constrain generalizability in data-sparse domains. Second, the proposed LAF-based aggregation and UTTL-based optimization methods may involve additional computational overhead due to multi-target representation and aggregation operations. Third, while the framework provides strong interpretability in uncertainty reasoning, its performance may vary depending on the alignment between the MIATTs structure and the underlying target distribution, requiring careful calibration of aggregation and weighting parameters. Finally, the current work focuses primarily on non-specific task settings; extension to classification, segmentation, sequential, generative, or reinforcement learning scenarios remains an open challenge.

Future research will extend EL-MIATTs along several promising directions: (1) Adaptive MIATT Generation and Weighting: develop self-adaptive mechanisms for generating and weighting MIATTs based on dynamic task characteristics or model uncertainty; (2) Paraconsistent and Multi-valued Reasoning: integrate paraconsistent logic and higher-order multi-valued semantics into LAF to better handle conflicting or contradictory MIATTs; (3) Scalable Implementation: explore efficient computational approximations or distributed architectures to reduce the cost of MIATT-based aggregation and optimization; and (4) Cross-domain and Real-world Proof-of-Concept: validate EL-MIATTs in broader real-world contexts, such as medical imaging, autonomous systems, and human-in-the-loop AI, where uncertainty and partial truth are intrinsic.

In summary, this work lays a theoretical and algorithmic foundation for bridging logical reasoning and statistical learning in uncertainty-aware machine learning. Future advancements in EL-MIATTs are expected to further strengthen its capacity to support robust, interpretable, and epistemically grounded AI systems. An application of this work's results is presented as part of the study available at https://www.qeios.com/read/EZWLSN.